\pdfoutput=1
\documentclass[11pt]{article}

\usepackage[]{ACL2023}

\usepackage{times}
\usepackage{latexsym}

\usepackage[T1]{fontenc}
\usepackage[utf8]{inputenc}

\usepackage{microtype}

\usepackage{inconsolata}

\usepackage{amsfonts} % blackboard字体
\usepackage{amssymb} % 数学符号
\usepackage{pifont}% Pi fonts
\usepackage{tabularx} % 表格长句自动换行
\usepackage{arydshln} % 加虚线 hdashline / cdashline
\usepackage{xcolor} % 颜色
\usepackage{mathtools} % 公式
\usepackage{adjustbox} % 表格的大小调整
\usepackage{multirow} % 表格中的跨行
\usepackage{paralist} % 提供了行距比较小的enumerate和itemize环境
\usepackage{CJKutf8} % 中文支持
\usepackage{booktabs} % 三线表
\usepackage{enumerate}
\usepackage{bbding} 
\usepackage{wasysym}
\usepackage{caption}
\usepackage{subcaption}

\usepackage[ruled,linesnumbered]{algorithm2e}
\usepackage{pythonhighlight}
\SetKwInOut{Parameter}{Parameters}

\usepackage{amsthm} % Theorems and proofs https://www.overleaf.com/learn/latex/Theorems_and_proofs
\theoremstyle{definition} % 直体
\usepackage{graphicx}
\usepackage{pythonhighlight}
\usepackage{listings}
\lstnewenvironment{mypython}[1][]{\lstset{style=mypython,#1}}{}
\usepackage{tikz}
\usepackage[edges]{forest} % forest流程图，直角连线
\usepackage{pgfplots}
\usepackage{custom} % 引入自定义的命令
\usepackage{amsmath}
\usepackage{bm}
\usepackage{scalefnt}

\newcommand{\openslu}[0]{\texttt{OpenSLU}}

\title{
	\openslu{}: A Unified, Modularized, and Extensible Toolkit \\ for Spoken Language Understanding
}

\author{
	Libo Qin$^{1}$\thanks{\ \ Equal Contribution}, Qiguang Chen$^{2}$\footnotemark[1], Xiao Xu$^{2}$, Yunlong Feng$^{2}$, Wanxiang Che$^{2}$ \\
	$^{1}$Central South University\\
	$^{2}$Research Center for Social Computing and Information Retrieval \\
	Harbin Institute of Technology, China \\
	{\texttt{\{lbqin,qgchen,xxu,ylfeng,car\}@ir.hit.edu.cn}}}	

\begin{document}
	\maketitle
	
	\begin{abstract}
Spoken Language Understanding (SLU) is one of the core components of a task-oriented dialogue system, which aims to extract the semantic meaning of user queries (e.g., intents and slots).
In this work, we introduce \openslu{}, an
open-source toolkit to provide a unified, modularized, and extensible toolkit for spoken language understanding. 
Specifically, \openslu{} unifies 10 SLU models for both single-intent and multi-intent scenarios, which support both non-pretrained and pretrained models simultaneously. 
Additionally, \openslu{} is highly modularized and extensible by decomposing the model architecture, inference, and learning process into reusable modules, which allows researchers to quickly set up SLU experiments with highly flexible configurations.
 \openslu{} is implemented based on PyTorch, and released at \url{https://github.com/LightChen233/OpenSLU}.
\end{abstract}
	\section{Introduction}
\label{Introduction}
\begin{figure}[t]
	\centering
	\centering
	\includegraphics[width=0.48\textwidth]{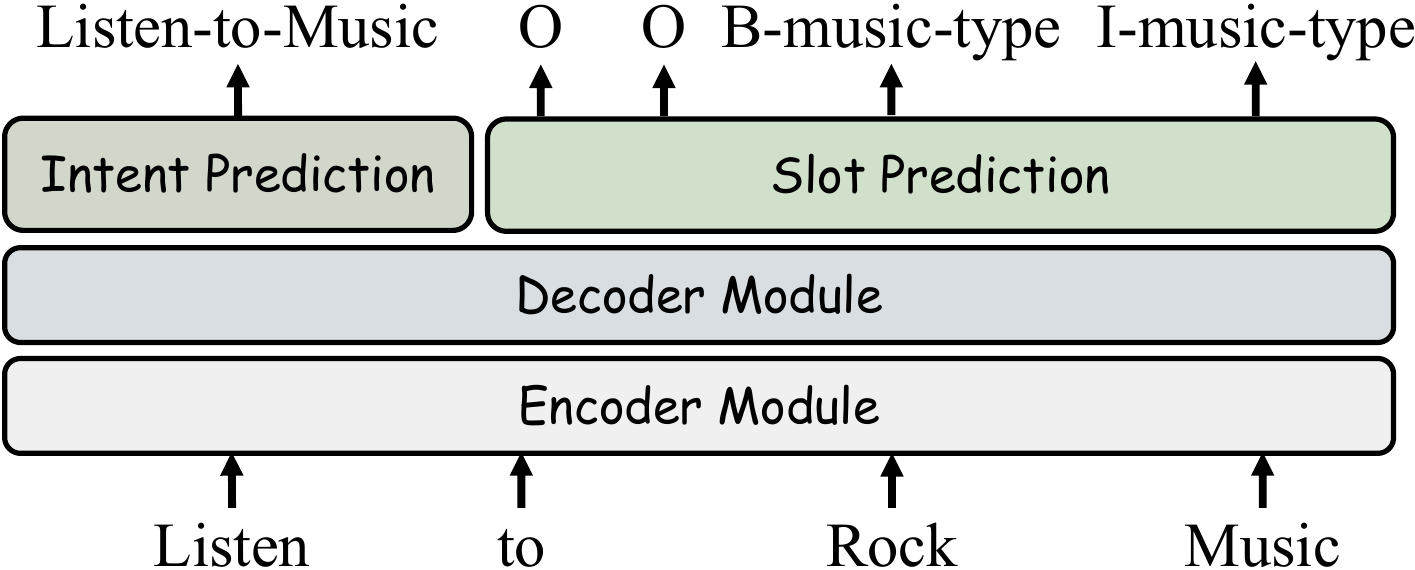}
	\caption{An example of spoken language understanding. \texttt{Listen-to-Music} stands for the intent label while \{O, O, B-music-type, I-music-type\} denotes the slot sequence labels.
	}
	\label{fig:intro}
\end{figure}

Spoken Language Understanding (SLU), which is used to extract the semantic frame of user queries (e.g., intents and slots)~\citep{tur2011spoken}. 
Typically, SLU consists of two sub-tasks: intent detection and slot filling. Take the utterance shown in Figure \ref{fig:intro} as an example, given ``\textit{Listen to Rock Music}'', the outputs include an intent class label (i.e., \texttt{Listen-to-Music}) and a slot label sequence (i.e., O, O, {B-music-type, I-music-type}).

Since intent detection and slot filling are highly tied~\citep{qin2021survey}, dominant methods in the literature explore joint models for SLU to capture shared knowledge~\cite{goo2018slot,wang-etal-2018-bi,qin-etal-2019-stack}.
Recently, \citet{gangadharaiah-narayanaswamy-2019-joint} shows that, in the amazon internal dataset, 52\% of examples contain multiple intents. Inspired by this observation, various SLU work shift their eye from single-intent SLU to multi-intent SLU scenario~\cite{gangadharaiah-narayanaswamy-2019-joint,qin2020agif,casanueva-etal-2022-nlu,moghe2022multi3nlu++}.

\begin{figure*}[t]
	\centering
	\includegraphics[width=\textwidth]{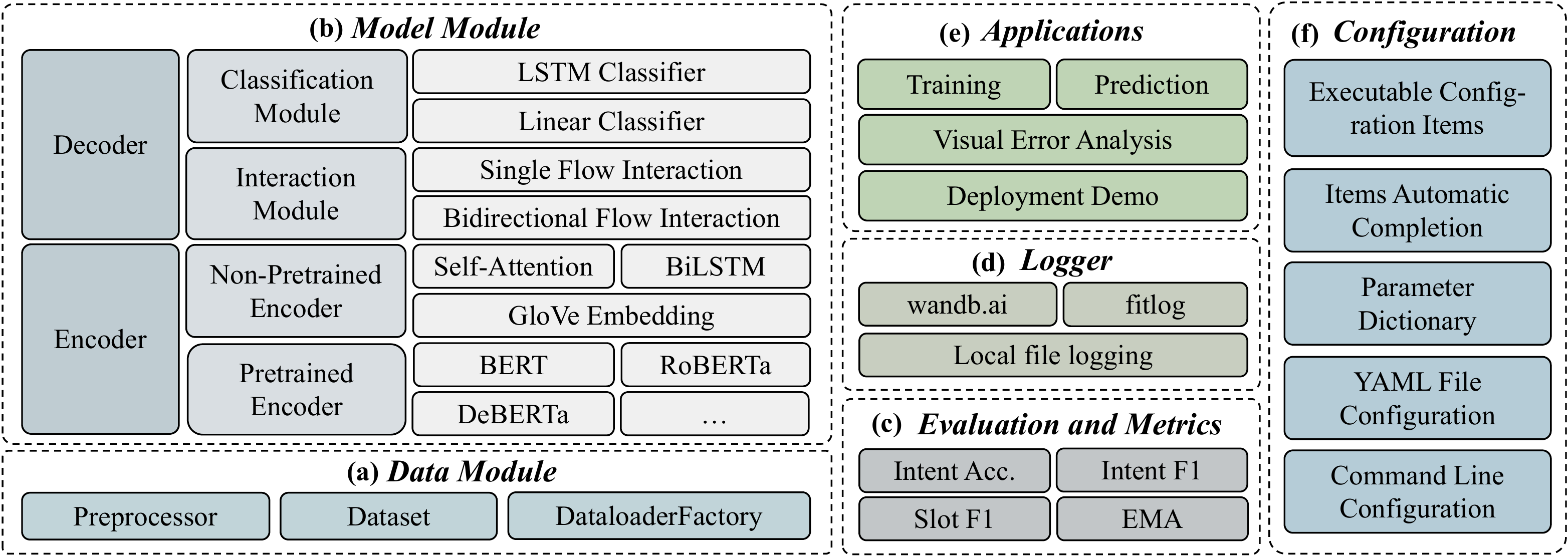}%
	\caption{Overall workflow of \openslu{}, which consists of (a) Data Module, (b) Model Module, (c) Evaluation and Metrics, (d) Logger, (e) Applications and (f) Configuration. }
	\label{fig:overall}
\end{figure*}

Thanks to the development of neural network, especially the successful use of large pretrained models, remarkable success have been witnessed in SLU. 
Nevertheless, there still lacks a unified open-source framework to facilitate the SLU community.
In this work, we make the first attempt to introduce \openslu{}, a unified, modularized, and extensible toolkit for SLU, which aims to help researchers to set up experiments and develop their new models quickly. 
The main features of \openslu{} are:

\begin{itemize}
\item \textbf{Unified and modularized toolkit}.
\openslu{} is the first unified toolkit to support both single-intent and multi-intent SLU scenarios.
Meanwhile, it is highly modularized by  
decoupling SLU models into a set of highly reusable modules, including data module, model module, evaluation module, as well as various common components and functions. 
Such modularization allows users to quickly re-implement SLU baselines or develop their new SLU models by re-using provided modules or adding new modules.

\item \textbf{Extensible and flexible toolkit}. 
\openslu{} is configured by configuration objects, which is extensible and can be initialized from YAML files. This enables users can easily develop their models by simply extending the configurations.
Additionally, we provide various interfaces of various common functions or modules in SLU models, including Encoder and Decoder module. 
Besides, the interfaces of our toolkit are fully compatible with the PyTorch interface, which allows seamless integration and flexibly rewriting any sub-module in the toolkit.

\item \textbf{Visualization Tool.}  
We provide a visualization tool to help users to view all errors of the model directly. 
With the help of visualization tool,  we can get a clearer picture: where we are and where we should focus our efforts to improve the performance of the model, which helps to develop a more superior framework.

\end{itemize}

To our knowledge, this is the first unified, modularized, and extensible toolkit for SLU.
We hope our work can help researchers to quickly initiate experiments and spur more breakthroughs in SLU\footnote{Video introduction about OpenSLU is available at \url{https://youtu.be/uOXh47m_xhU}.}.

\section{Architecture and Design}
\label{sec:dataset-construction}
Figure~\ref{fig:overall} illustrates the overall workflow of \openslu{}. In this section, we describe the (a) Data Module ($\S \ref{sec:data}$); (b) Model Module; ($\S \ref{sec:model}$); (c) Evaluation and Metrics ($\S \ref{sec:evaluation}$) and other common modules (Logger, Applications and Configuration module) ($\S \ref{sec:common}$).
\subsection{Data Module}\label{sec:data}
\openslu{} offers an integrated data format in the data module (see Figure~\ref{fig:overall}(a)) for SLU models, which can be denoted as:
$ \textit{raw text} \rightarrow \texttt{Preprocessor}  \rightarrow \texttt{Dataset} \rightarrow \texttt{DataLoaderFactory} \rightarrow \textit{model input}$. 

Given the input \textit{raw text}, $\textit{Preprocessor}$ sub-module first pre-process different raw texts to an integrated $.jsonl$ format that contains slot, text and intent, which is formatted as: 
\begin{python}
{
	"slot": [List of Slot Value],
	"text": [List of Text],
	"intent": [Intent Value]
}.
\end{python}

The $\texttt{Dataset}$ sub-module offers a range of data processing operations to support both pretrained and non-pretrained models. For pretrained models, these operations include lowercase conversion, BPE-tokenization, and slot alignment, while for non-pretrained models, the sub-module handles word-tokenization and vocabulary construction. 

Finally, \texttt{DataLoaderFactory} sub-model is used for creating \texttt{DataLoader} to manage the data stream for models.

\subsection{Model Module}\label{sec:model}
As shown in Figure~\ref{fig:overall}(b), the overall model module contains encoder module($\S \ref{sec:encoder}$) and decoder module ($\S \ref{sec:decoder}$).
\subsubsection{Encoder}\label{sec:encoder}
For the encoder module, we implement both non-pretrained models and pretrained models.
In non-pretrained models, we offer the widely used SLU encoders including self-attentive~\cite{vaswani2017attention, qin-etal-2019-stack} and BiLSTM~\cite{hochreiter1997long, goo2018slot, liu2020attention}  encoder.
Additionally, we support auto-load GloVe embedding \citep{pennington2014glove}.

In pretrained models, \openslu{} supports various encoders including BERT~\cite{devlin2019bert}, RoBERTa~\cite{liu2019roberta}, ELECTRA~\cite{clark2020electra}, DeBERTa$_{v3}$~\cite{he2020deberta}.

\subsubsection{Decoder}\label{sec:decoder}
Since slot filling and intent detection are highly related, dominant methods in the literature employ joint models to capture the shared knowledge across the related tasks~\cite{goo2018slot, wang-etal-2018-bi, chen2019bert}.
To support the joint modeling paradigm, 
decoder in \openslu{} contains two sub-modules: (1) interaction module for capturing interaction knowledge for slot filling and intent detection and (2) classification module for the final prediction results.

\begin{figure}[t]
	\centering
	\includegraphics[width=0.45\textwidth]{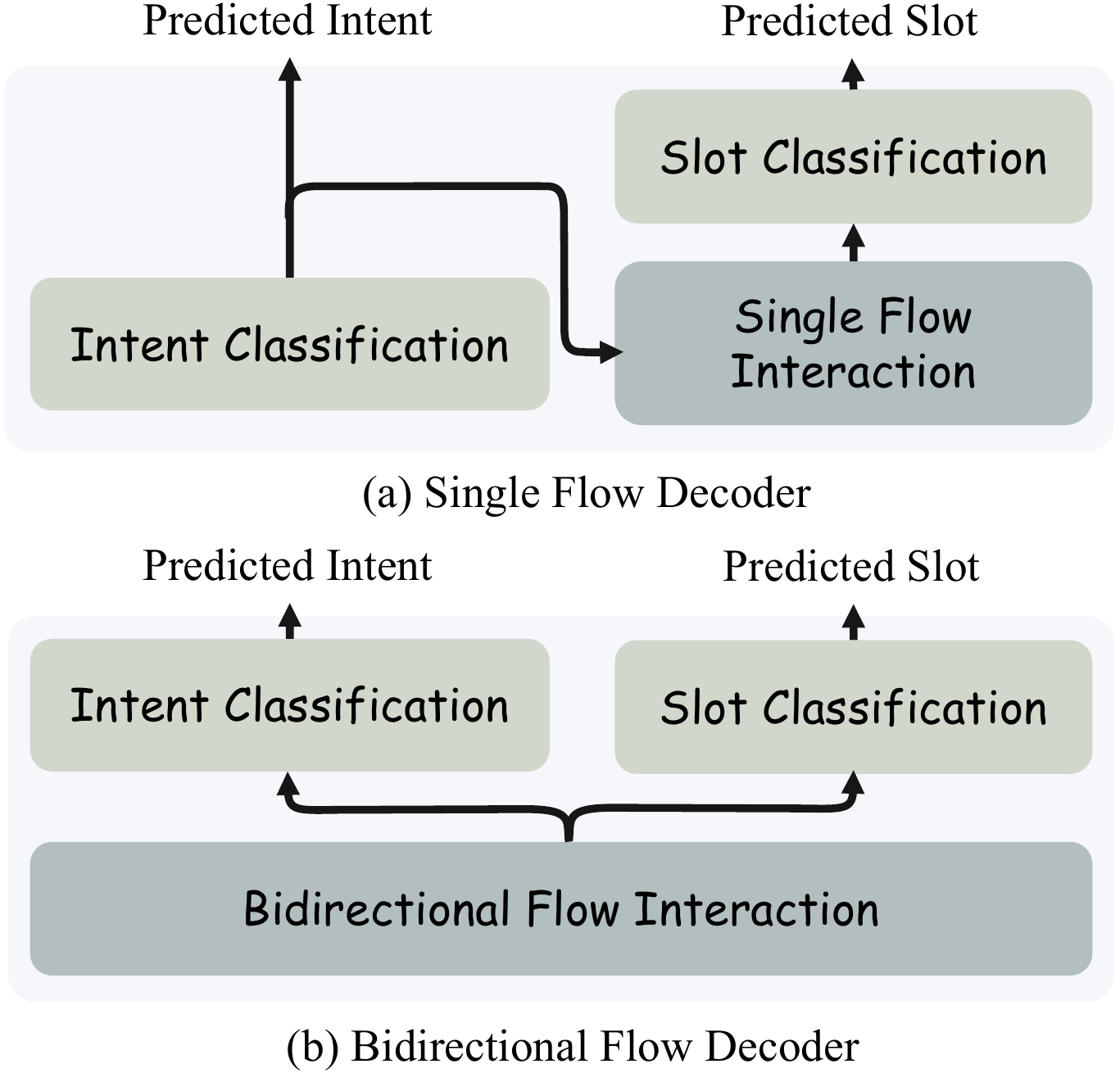}
	\caption{Single flow interaction decoder (a) vs. bidirectional flow interaction decoder (b).}
	\label{fig:decoder}
\end{figure}

\paragraph{Interaction Module.} 
As summarized in \citet{qin2021survey}, interaction module consists of two widely used the interaction types, including \textit{single flow interaction} and \textit{bidirectional flow interaction}.
\begin{itemize}
	\item \textbf{Single Flow Interaction} refers to the flow of information from intent to slot in one direction as illustrated in Figure~\ref{fig:decoder}(a). A series of studies~\cite{goo2018slot, li2018self, qin-etal-2019-stack}
	have achieved remarkable improvements in performance by guiding slot filling with intent detection information.
	\item \textbf{Bidirectional Flow Interaction}  stands for that the bidirectional cross-impact between intent detection and slot filling can be considered, which is shown in Figure~\ref{fig:decoder}(b). Another series of works~\cite{wang-etal-2018-bi, niu2019novel, liu2019cm, qin2021co} build the bidirectional connection across slot filling and intent detection to enhance each other.

\end{itemize}

 Based on the two types of interaction, users can easily design the interaction module and interaction order via our provided classic interaction modules and customized configurations.

\paragraph{Classification Module.}
It aims to transform hidden states after interaction module into final classification logits.
There are two types of classification modules supported by \openslu{}:
\begin{itemize}
		
			\item \textbf{MLP Classifier.} Multi-Layer Perceptron (MLP) Classifier is a fundamental classification decoding algorithm.
			Nevertheless, the method ignores the dependency across tokens.

		\item \textbf{LSTM Classifier.}  
		It indicates that we adopt an LSTM classifier for the final prediction, which has the advantage of modeling the dependency of tokens (from left to right). 
		However, it is an autogressive classification module for SLU, which cannot be parallel to speed up the decoding prediction.

\end{itemize}

To improve the quality of SLU prediction results, we also implement several SLU tricks, like 
 teacher-forcing and token-level intent detection~\cite{qin-etal-2019-stack}.
 Users can switch between different prediction strategies by simply setting up the hyper-parameter to improve performance.

\begin{figure*}[t]
	\centering
	\includegraphics[width=\textwidth]{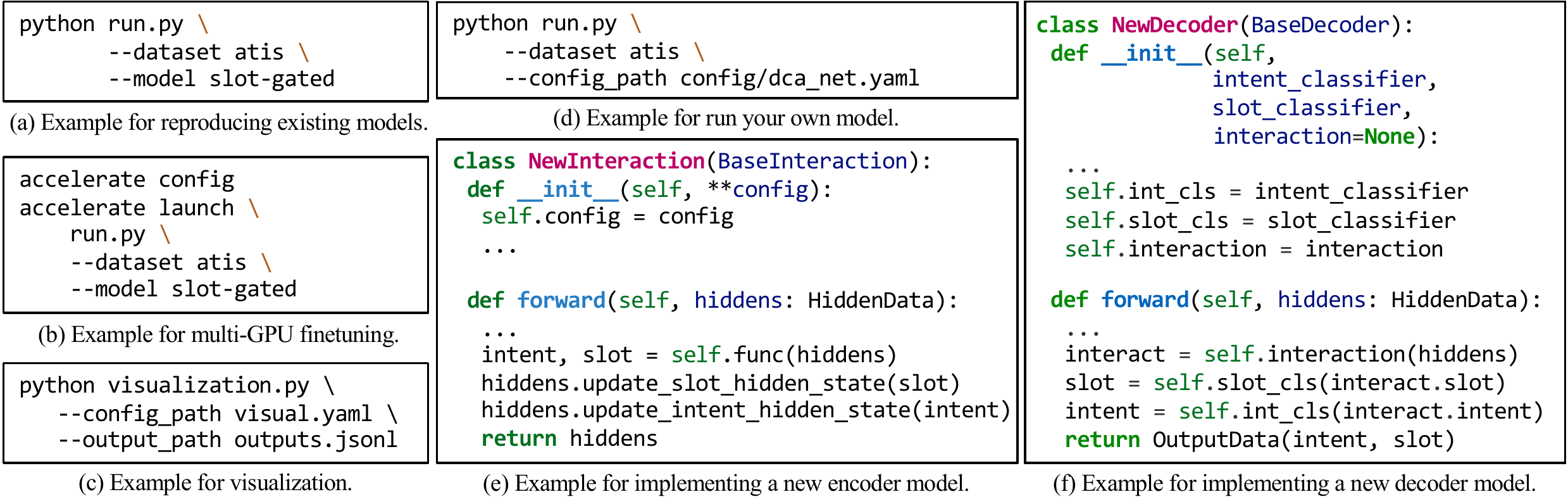}%
	\caption{ Example usage of \openslu{}.
	}
	\label{fig:code}
\end{figure*}
\subsection{Evaluation and Metrics}\label{sec:evaluation}
Following \citet{goo2018slot,qin2021survey}, we support various metrics for SLU (shown in Figure~\ref{fig:overall}(c)), including Slot F1 Score, Intent Accuracy, Intent F1, and Exactly Match Accuracy (EMA).

\begin{itemize}
	\item \textbf{Slot F1 Score}~\citep{goo2018slot, qin-etal-2019-stack} is used for assessing slot filling performance. This metric is calculated as the harmonic mean between precision and recall. 
	\item \textbf{Intent Accuracy}~\citep{goo2018slot, qin-etal-2019-stack} is a measure used to evaluate the accuracy of intent detection, based on the ratio of correctly predicted intents.
	\item \textbf{Intent F1 Score}~\citep{gangadharaiah-narayanaswamy-2019-joint, qin2020agif} is adopted to evaluate the macro F1 Score of the predicted intents in the multi-intent detection.
	\item \textbf{Exact Match Accuracy}~\citep{goo2018slot, qin-etal-2019-stack, qin2020agif} takes intent detection as well as slot filling into account simultaneously. This metric is calculated as the ratio of sentences for which both the intent and slot are predicted correctly within a sentence.
\end{itemize}

\subsection{Common Modules}\label{sec:common}
\paragraph{Logger.} We provide a generic \texttt{Logger} component to help users to track the process of model building including wandb.ai, fitlog and local file logging (see Figure~\ref{fig:overall}(d)). 
\paragraph{Applications.} 
We provide complete scripts in the \texttt{Application} (see Figure~\ref{fig:overall}(e)) for training, prediction, visual error analysis, and the final stage of model deployment. 
\paragraph{Configuration.} 
As shown in Figure~\ref{fig:overall}(f), our toolkit employs \texttt{Configuration} module to manage the model configuration, training parameters, and training and analysis data. We will introduce more details in Section Toolkit Usage ($\S \ref{sec:usage}$).

	\section{Toolkit Usage}\label{sec:usage}

\begin{table*}
	%	\vspace{-2mm}
	\centering
	\begin{adjustbox}{width=\textwidth}
		\begin{tabular}{l|ccc|ccc}
			\hline
			\multicolumn{1}{c|}{\multirow{2}{*}{Model}} &\multicolumn{3}{c|}{ATIS}&\multicolumn{3}{c}{SNIPS}\\
			\cline{2-7}
			&Slot F1.(\%)& Intent Acc.(\%) & EMA(\%)& Slot F1.(\%) & Intent Acc.(\%) & EMA(\%)\\
			\hline
			\multicolumn{7}{c}{\textit{Non-Pretrained Models}} \\
			\hline
			Slot Gated~\cite{goo2018slot}& 94.7 & 94.5 & 82.5 & 93.2 & 97.6 & 85.1\\
			Bi-Model~\cite{wang-etal-2018-bi}& 95.2 & 96.2 & 85.6 &  93.1 & 97.6  & 84.1 \\
			Stack Propagation~\cite{qin-etal-2019-stack}& 95.4 & 96.9 & 85.9 & \textbf{94.6} & 97.9 & 87.1 \\
			DCA Net~\cite{qin2021co}& \textbf{95.9} & \textbf{97.3} & \textbf{87.6} & 94.3 & \textbf{98.1} & \textbf{87.3}\\
			
			\hline
			\multicolumn{7}{c}{\textit{Pretrained Models}} \\
			\hline
			Joint BERT~\cite{chen2019bert}& \textbf{95.8} & \textbf{97.9} & \textbf{88.6} & 96.4 & 98.4 & 91.9 \\
			RoBERTa~\cite{liu2019roberta} & 95.8 & 97.8 & 88.1 & 95.7 & 98.1 & 90.6\\
			ELECTRA~\cite{clark2020electra} & 95.8 & 96.9 & 87.1 & 95.7 & 98.3 & 90.1 \\
			DeBERTa$_{v3}$~\cite{he2020deberta} & 95.8 & 97.8 & 88.4 & \textbf{97.0} & \textbf{98.4} & \textbf{92.7} \\
			\hline
		\end{tabular}
	\end{adjustbox}
	\caption{
		Single-intent SLU main Results. All baseline results are re-implemented by \openslu{}. 
	}
	\label{exp:main_results}
\end{table*}

\subsection{Reproducing Existing Models}
For reproducing an existing model implemented by \openslu{} on different datasets, users are required only to specify the dataset and model by setting hyper-parameters, i.e., $\textit{model}$ and $\textit{dataset}$. 
Experiments can be reproduced in a simple command line instruction, as shown in Figure~\ref{fig:code}(a).
This instruction aims to fine-tuning \texttt{Slot-Gated}~\cite{goo2018slot} model on ATIS~\cite{hemphill1990atis} dataset. 
With YAML configuration files, we can modify hyper-parameters conveniently, which allows users can reproduce various experiments quickly without modifying the source code. 
In addition, we designed \openslu{} to work on a variety of hardware platforms.
If the hyper-parameter $\textit{device}$ is set to ``\textit{cuda}'', CUDA devices will be used. Otherwise, CPU will be employed by default. 
As shown in Figure~\ref{fig:code}(b), we also support distributed training on multi-GPU by setting hyper-parameters and command line parameters.

\subsection{Customizable Combination Existing Components}
As the model is designed as reusable modules, users can easily reuse modules via the call of interface or configuration files.
More specifically, for the interface, users can call common-used encoder and decoder modules in one line of code from the pre-configured library.
For configuration files, users can combine existing component libraries only through configuration files, thus creating a customized model. 

It can be useful for users in cross-cutting areas, such as biology, that are unfamiliar with using Python code to create models, as it allows them to create their own models without using any Python code.
Such feature can potentially make it easier to build and test models more rapidly. Similarly, the customized model can be trained by specifying the relevant configuration file path and running simple command line instructions, as shown in Figure~\ref{fig:code}(d).
\begin{figure}[t]
		\centering
		\resizebox{\columnwidth}{!}{%
			\begin{tikzpicture} 
				\begin{axis}[
					enlargelimits=0.5,
					legend style={at={(0.5,-0.25)},
						anchor=north,legend columns=-1},
					ylabel={EMA.(\%)},
					ylabel near ticks,
					xlabel near ticks,
					ymajorgrids=true,
					grid style=dashed,
					symbolic x coords={MixATIS,MixSNIPS},
					xtick=data,
					ybar=8pt,
					bar width=0.5cm,
					nodes near coords,
					nodes near coords align={vertical},
					nodes near coords style={font=\tiny},
					font=\small,
					grid=major,
					width=\linewidth,
					ytick align=inside,
					xtick align=inside,
					height=0.45\linewidth,
					]

					\addplot [fill=brown!40!white,draw=brown]
					coordinates {(MixATIS,36.4) (MixSNIPS,70.2)};
					
					\addplot [fill=black!40!white,draw=black]
					coordinates {(MixATIS,39.5)  (MixSNIPS,74.8)};

					\addplot [fill=blue!40!white,draw=blue]
					coordinates {(MixATIS,42.4) (MixSNIPS,74.9)};
					\legend{Multi-task,AGIF,GL-GIN}
				\end{axis}
		\end{tikzpicture}}
\caption{Multi-intent SLU main results on EMA. All baseline results are re-implemented by \openslu{}.}
\label{fig:multi}
\end{figure}
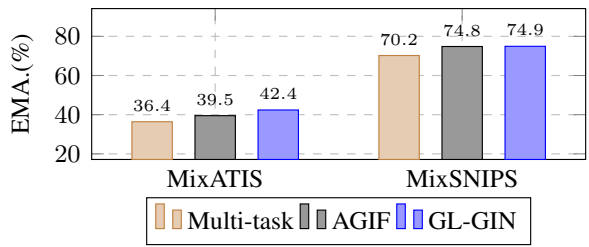
\subsection{Implementing a New SLU Model}
\label{sec:CLIP}
Since \openslu{} split the model into fine-grained components, users can directly reuse modules through configuration files.
Specifically, when users aim to implement a new SLU model, only a few key innovative modules need to be rewritten by users, including 
a specific Model class and 2 functions as follows:
  
\noindent $\bullet$ \textit{\_\_init\_\_()} function. This function aims for parameter initialization, global variable definition, and so on.
All modules can be inserted into the system by configuring the \textit{\_\_model\_target\_\_} hyper-parameters, so as to quickly and automatically build the model.

\noindent $\bullet$ \textit{forward()} function. This function mainly focuses on forward data flow and learning the parameters according to the pre-defined configuration.

In most cases, rewriting \texttt{Interaction} module is enough for building a new SLU model. As shown in Figure~\ref{fig:code}(e), this module accepts \texttt{HiddenData} data object as input and return with \texttt{HiddenData} data object. \texttt{HiddenData} contains the \textit{hidden\_states} for intent and slot, and other helpful information.
With the advancement of SLU research, patterns of decoders become increasingly complex~\citep{xing2022co, cheng2022scope}. Therefore, to further meet the needs of complex exploration, we provide the \texttt{BaseDecoder} class, and the user can simply override the \textit{forward()} function in class, which accepts $\texttt{HiddenData}$ as input data format and $\texttt{OutputData}$ as output data format, as shown in Figure~\ref{fig:code}(f).

\begin{figure*}[t]
	\centering
	\includegraphics[width=0.95\textwidth]{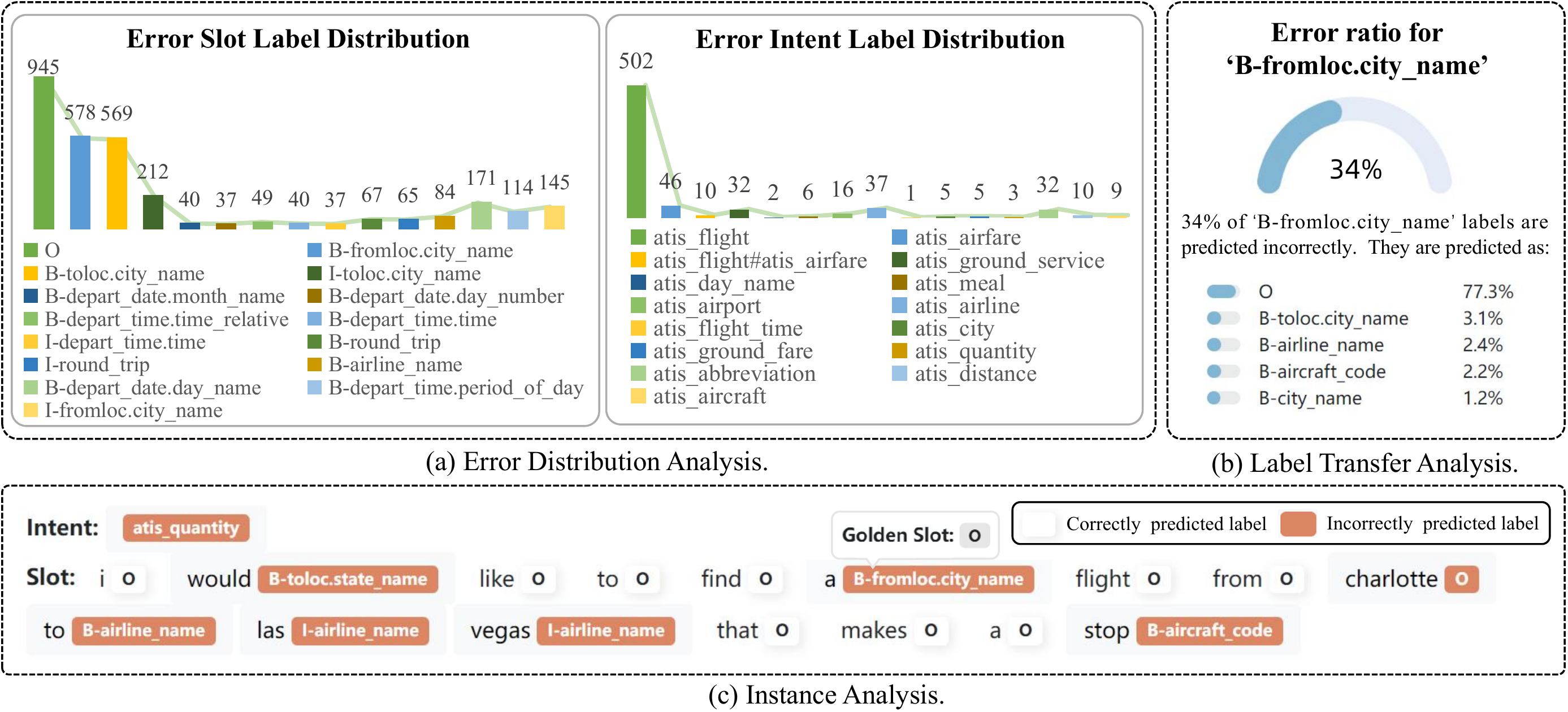}%
	\caption{Visualization Analysis. It mainly contains three functions including Error Distribution Analysis (a), Label Transfer Analysis (b) and Instance Analysis (c).
	}
	\label{fig:demo}
\end{figure*}

\section{Experiments}
\label{experiments}
Extensive reproduction experiments are conducted to evaluate the effectiveness of \openslu{}.
\subsection{Data Settings}
In single-intent SLU, we employ two widely used benchmarks including ATIS~\citep{hemphill1990atis} and SNIPS dataset~\cite{coucke2018snips}.

In multi-intent SLU scenario, we support 2 widely used datasets: MixATIS and MixSNPIS~\cite{qin2020agif}, which are collected from the ATIS, SNIPS by simple conjunctions, e.g., ``and'', to connect sentences with different intents. 

\subsection{Result Reproduction}

We implement various state-of-the-art SLU models.
For single-intent SLU methods, we re-implement the following baselines:
(1) \texttt{Slot Gated}~\cite{goo2018slot};
(2) \texttt{Bi-Model}~\cite{wang-etal-2018-bi};
(3) \texttt{Stack Propagation}~\cite{qin-etal-2019-stack};
(4) \texttt{DCA Net}~\cite{qin2021co};
(5) \texttt{Joint Bert}~\cite{chen2019bert};
(6) \texttt{RoBERTa}~\cite{liu2019roberta};
(7) \texttt{ELECTRA}~\cite{clark2020electra};
(8) \texttt{DeBERTa$_{v3}$}~\cite{he2020deberta}.
For multi-intent SLU methods, we adopt the following baselines:
(1) \texttt{AGIF}~\cite{qin2020agif};
(2) \texttt{GL-GIN}~\cite{qin2021gl}.

The reproduction results are illustrated in Table~\ref{exp:main_results}, we observe that \openslu{} toolkit can reproduce the comparable results reported in previous works, which verify the effectiveness of \openslu{}.
In addition, \openslu{} can outperform some reported results in previous published work, which further shows the superiority of \openslu{}.
Meanwhile, the same trend can be observed in multi-intent SLU setting, which is shown in Figure~\ref{fig:multi}.

\subsection{Visualization Analysis}\label{sec:visual}
According to a number of studies~\citep{vilar2006error,wu2019errudite, ribeiro2020beyond, paleyes2022challenges}, model metrics tests alone no longer adequately reflect the model's performance. To help researchers further improve their models, we provide a tool for visual error analysis including three main parts: (a) error distribution analysis; (b) label transfer analysis; and (c) instance analysis (see Figure~\ref{fig:demo}).
And the visual analysis interface can be run with the command as shown in the Figure~\ref{fig:code}(c).

\subsubsection{Error Distribution Analysis.} 
We provide error distribution analysis that presents the number and percentage of label errors predicted by the model.
By viewing the error distributions, the model can be easily analyzed and studied qualitatively~\citep{caubriere2020error}. As a result, the weaknesses of each system can be better understood and improvements can be made to the model in the future.

Take the error in Figure~\ref{fig:demo}(a) as an example, 
a large number of \textit{atis\_flight} labels are incorrectly predicted compared with all other labels. Therefore, we should pay more attention on how to improve the performance of \textit{atis\_flight} labels.

\subsubsection{Label Transfer Analysis.} 
Label Transfer Analysis module first offers the percentage of incorrect predictions for each label and provides the probability of being incorrectly predicted as each of the other labels to present a fine-grained statistics for a better understanding of issues such as invisible bias in the model~\citep{wu2019errudite, ribeiro2020beyond}.

For example, Figure~\ref{fig:demo}(b) shows the details in incorrect prediction on `\textit{B-fromloc.city\_name}'. We  observe 34\% of `\textit{B-fromloc.city\_name}' predict incorrectly and 77.3\% of error labels are predicted as `\textit{O}'. By having access to this information, users can be better guided to improve their data or label learning methods to prevent those error predictions.

\subsubsection{Instance Analysis.} 
In order to provide a better case study, \openslu{} offers a instance-level analysis view by highlighting error results and interactively checking all golden labels (shown in Figure~\ref{fig:demo}(c)).
 Such instance analysis allows users to examine data on a case-by-case basis in an intuitive way.
This can be seen easily in  Figure~\ref{fig:demo}(c), where token `\textit{a}' is predicted as `\textit{B-fromloc.city\_name}' instead of `\textit{O}'.

Furthermore, we also deploy \openslu{} into the Gradio\footnote{https://www.gradio.app} platform, which allows users to connect the demo directly to the public network and access it via the computer or mobile device.
	\section{Conclusion}
\label{conclusion}
This paper introduce \openslu{}, a unified, modularized, and extensible toolkit for spoken language understanding. In our toolkit, we implement 10 models on both single-intent setting and multi-intent SLU settings, both covering the categories of non-pretrained and pretrained language models. 
Our toolkit is plug-and-play and can be easily applied to other SLU setting, which is extensible to support seamless
incorporation of other external modules.
To the best of our knowledge, this is the first open-resource toolkit for SLU and we hope \openslu{} can attract more breakthroughs in SLU. In the future, we can extend \openslu{} to support cross-lingual scenario~\cite{ijcai2020p533,zheng-etal-2022-hit}.

\section*{Acknowledgements}
This work was supported by the National Key R\&D Program of China via grant 2020AAA0106501 and the National Natural Science Foundation of China (NSFC) via grant 62236004 and 61976072.
	
	\bibliography{anthology,custom}
	\bibliographystyle{acl_natbib}
	
	\appendix
	
	\clearpage
\end{document}